\documentclass[10pt,conference]{IEEEtran}
\IEEEoverridecommandlockouts
\pdfoutput=1
\usepackage{cite}
\usepackage{amsmath,amssymb,amsfonts}
\usepackage{algorithmic}
\usepackage{graphicx}
\usepackage{textcomp}
\usepackage{xcolor}
\usepackage[numbers,sort]{natbib}
\usepackage{xcolor}
\usepackage{algorithmic}
\usepackage{graphicx}
\usepackage{textcomp}
\usepackage{xspace}
\usepackage{hyperref}
\usepackage{url}

\usepackage[nameinlink, capitalize]{cleveref}
\usepackage{tabularx, booktabs}
\usepackage{multirow}
\usepackage{caption}
\usepackage{subcaption}
\usepackage{float}
\usepackage{framed}
\usepackage{multicol}

\def\BibTeX{{\rm B\kern-.05em{\sc i\kern-.025em b}\kern-.08em
		T\kern-.1667em\lower.7ex\hbox{E}\kern-.125emX}}
	
\definecolor{linkcolor}{rgb}{0.65,0,0}
\definecolor{citecolor}{rgb}{0,0.4,0}
\definecolor{urlcolor}{rgb}{0,0,0.65}
\definecolor{TolDarkGreen}{HTML}{117733}

\hypersetup{hyperindex=true,pdfpagemode=UseNone,pdfstartview=FitH,colorlinks=true, linkcolor=linkcolor, urlcolor=urlcolor, citecolor=citecolor}
\newcolumntype{C}[1]{>{\centering\arraybackslash}m{#1}}

\newcommand{\parhead}[1]{\vspace{1pt plus 0.5pt minus 0.5pt}\par\noindent\textbf{#1}\hspace{.75em plus .5em minus .5em}}
\crefname{figure}{Figure}{Figures}

\def\BibTeX{{\rm B\kern-.05em{\sc i\kern-.025em b}\kern-.08em
    T\kern-.1667em\lower.7ex\hbox{E}\kern-.125emX}}

\begin{document}

\title{Revisiting Lightweight Compiler Provenance Recovery on ARM Binaries}

\author{\IEEEauthorblockN{Jason Kim}
\IEEEauthorblockA{Georgia Tech\\
nosajmik@gatech.edu}
\and
\IEEEauthorblockN{Daniel Genkin}
\IEEEauthorblockA{Georgia Tech\\
genkin@gatech.edu
}
\and
\IEEEauthorblockN{Kevin Leach}
\IEEEauthorblockA{Vanderbilt University\\
kevin.leach@vanderbilt.edu}
}

\maketitle

\begin{abstract}
A binary's behavior is greatly influenced by how the compiler builds its source code.
Although most compiler configuration details are abstracted away during compilation, recovering them is 
useful for reverse engineering and program comprehension tasks on unknown binaries, such as code similarity detection.
We observe that previous work has thoroughly explored this on x86-64 binaries. 
However, there has been limited investigation of ARM binaries, which are increasingly prevalent.

In this paper, we extend previous work with a shallow-learning model that efficiently and accurately recovers compiler configuration properties for ARM binaries. 
We apply opcode and register-derived features, that have previously been effective on x86-64 binaries, to ARM binaries. 
Furthermore, we compare this work with Pizzolotto et al., a recent architecture-agnostic model that uses deep learning, whose dataset and code are available.
 
We observe that the lightweight features are reproducible on ARM binaries. We achieve over 99\% accuracy, on par with state-of-the-art deep learning approaches, while achieving a 583-times speedup during training and 3,826-times speedup during inference. 
Finally, we also discuss findings of overfitting that was previously undetected in prior work. 
\end{abstract}


\section{Introduction} \label{s:intro}
The \emph{compiler provenance} (a.k.a. toolchain provenance) of a binary is
most commonly defined as the family, version, and optimization level of the
compiler used to produce the binary~\cite{Rosenblum10, Rosenblum11, Rahimian15,
Benoit21, Ji21, Massarelli19, Otsubo20}. While the compiler provenance of many
commercial off-the-shelf (COTS) binaries is not explicitly available due to
abstraction during compilation or intentional obfuscation, 
recovering this information is important in reverse engineering and program
comprehension tasks that involve malicious, obfuscated, or legacy binaries.

For example, compilers may induce unintended side
effects when optimizing on certain pieces of
code~\cite{Wang13OptimizationSecurity}, leading to the omission of memory
scrubbing for sensitive data~\cite{Yang17DeadStore,
DSilva15OptimizationSecurity, Hohnka19CompilerVuln} and emergence of timing
differences in constant-time code~\cite{DSilva15OptimizationSecurity,
Hohnka19CompilerVuln}.
On another angle, Ji et al.~\cite{Ji21} use compiler provenance to compile the source
code in the most accurate environment possible as that of the binary, where
they can subsequently treat the problem as a similarity task between binaries.

Moreover, it is important for binaries in safety-critical
systems to be guaranteed they will function exactly as their 
source code specifies. For instance, Airbus uses
CompCert~\cite{Leroy16CompCert}, a formally verified C compiler, for its flight
control software~\cite{Francca11AirbusCompCert, Kastner17AirbusCompCert,
Souyris13AirbusCompCert}. That is, an accurate model that
recovers provenance from a certain compiler may assist in comprehending 
correctness guarantees for closed-source COTS binaries.

Existing work has primarily considered accuracy~\cite{Rosenblum10, Rosenblum11, Benoit21,
Chen18RNNOpti, Ji21, Otsubo20, Pizzolotto20, Pizzolotto21, Tian21}. However,
the growing volume of binaries that must be reverse engineered has also made throughput an important consideration~\cite{Goues12APR, Harman12AIinSE, Gazzola17APR, Polino15APR}.  
A fast model is beneficial not only for efficiency, but also for explainability: iteratively excluding subsets of training data and retraining can help developers understand the contribution of removed subsets to the model's behavior, and check for biases~\cite{bellamy2019ai, mehrabi2021survey, pradhan2021interpretable}.
Furthermore, understanding how the model makes decisions is crucial for debugging and combating overfitting~\cite{carvalho2019machine, gilpin2018explaining, krishnan2017palm, rauschmayr2021amazon}, creating a need for a model that is accurate, fast, and readily interpretable.

In line with recent machine learning trends forgoing shallow models for deep neural networks, many works use the latter to recover
compiler provenance~\cite{Chen18RNNOpti, Yang19CNNOpti,
Massarelli19, Otsubo20, Pizzolotto20, Benoit21, Ji21, Tian21, Pizzolotto21}.
However, neural networks are not a panacea for classification. Their complexity makes them expensive to train not only in time, but also energy~\cite{strubell2019energy, patterson2021carbon, lacoste2019quantifying, strubell2020energy}: the CO$_2$ footprint of training a large AI model can exceed 626,000 pounds~\cite{strubell2019energy}.
In addition, these models preclude straightforward interpretation.
As such, we desire compiler provenance classification without the costs associated with deep neural networks. 

Furthermore, a substantial body of work on such models apply only to x86 and x86-64 binaries~\cite{Rosenblum10, Rosenblum11, Rahimian15, Chen18RNNOpti, Massarelli19, Otsubo20, Pizzolotto20, Benoit21, Ji21, Tian21,dicomp}.
Despite the
growth of the ARM architecture in embedded, mobile, and recently desktop
computing~\cite{ARMPresentation, ArmMarketShare, SoftBankARM, Graviton,
32bitMarketShare}, limited work has explored compiler provenance for ARM binaries~\cite{Yang19CNNOpti, Pizzolotto21, BinProv}. 

With this in mind, we adapt the most recent state-of-the-art work in compiler provenance for x86-64 binaries, called DIComP~\cite{dicomp}, to apply to 32- and 64-bit ARM binaries. 
DIComP uses a shallow learning model to recover compiler provenance information, but does not consider ARM. 
Thus, we discuss how we adapted their approach to work with ARM binaries. 
Furthermore, we compare against a state-of-the-art deep learning approach for compiler provenance, described in Pizzolotto et al.~\cite{Pizzolotto21}. 
Pizzolotto et al. describe an architecture-agnostic approach to compiler provenance recovery, and they share a large dataset of binaries and model source code.
 
We compare our approach to compiler provenance recovery against Pizzolotto's deep learning approach using their benchmark. 
We show that our shallow-learning approach can achieve classification performance for ARM binaries that is on par with state-of-the-art deep learning methods, while using a fraction of the time and resources. 

\parhead{Our Contributions.}

In this paper, we augment two previously published studies: DIComP~\cite{dicomp} and Pizzolotto et al.~\cite{Pizzolotto21}.
DIComP provides a shallow learning approach to recover compiler provenance information from x86-64 binaries, but not ARM.
On the other hand, Pizzolotto provides a deep learning approach that is architecture-agnostic.
We reimplement techniques described in DIComP to provide support for 32- and 64-bit ARM binaries, and then evaluate these techniques with respect to model accuracy, scalability, and interpretability.

Based upon our investigation of corpora of ARM binaries, we adapt a lightweight set of features that
aggregates ARM assembly into a representation emphasizing how the compiler uses
general-purpose registers and uncommon opcodes. 
We build a hierarchy of linear support vector machines (SVMs) that quickly and accurately recovers provenance information from ARM binaries.
We leverage SVMs for their fast training and evaluation speed, and direct correlation of classifier weights to features (unlike other shallow models such as random forests or boosting).

We present an evaluation that directly compares accuracy and runtime performance with the shallow-learning techniques shown in DIComP~\cite{dicomp} against the deep-learning techniques described in Pizzolotto et al.~\cite{Pizzolotto21}.
Here, we reproduce similar accuracy on ARM binaries compared to DIComP and Pizzolotto.
We also show that our model can be trained 583 times faster,
and can classify samples 3,826 times faster than state-of-the-art deep learning approaches.
Akin to DIComP, this speedup makes training and evaluation of large binary corpora feasible even without dedicated GPU hardware.

Subsequently, we demonstrate
this performance is consistent on 32-bit ARM binaries, and also achieve a near-perfect
accuracy in distinguishing certified CompCert binaries from `conventional' GCC or Clang binaries. 
In addition, we leverage the interpretability of our model to explore various hypotheses for differences in compiler backends that lead to these discernible differences among compiler provenances, mirroring DIComP's analysis for x86-64 on ARM. 
Finally, we address an overfitting issue in the dataset of Pizzolotto et al., and contribute the most comprehensive comparison of works on compiler
provenance recovery to our knowledge.

This replication study provides a valuable approach to generalizing results found in DIComP to support ARM binaries.
We release prototypes of our tool on GitHub\footnote{\url{https://github.com/lcp-authors/lightweight-compiler-provenance}} and Zenodo\footnote{\url{https://zenodo.org/record/6361049}}.

\begin{table*}[htb]
	\centering
	\def\arraystretch{1.5}
	\caption{Summary of preceding works that recover the compiler provenance of binaries with a machine learning model.\label{f:SoK}}
	\footnotesize
	\begin{tabular}{l|lllllllllll}
		\hline
		\textbf{Work} & \textbf{Year} & \textbf{Arch(s).} & \textbf{NN?$*$} & \textbf{Model Type} & \textbf{Attn.$\dagger$} & \textbf{CFG} & \textbf{Regs.$\ddagger$} & \textbf{Family$\mathsection$} & \textbf{Version$\mathsection$} & \textbf{Opti.$\mathsection$} & \textbf{Granularity} \\
		\hline
		Rosenblum~\cite{Rosenblum10} & 2010 & x86-64 & No & CRF & --- & No & Yes & 3 & --- & --- & Byte \\
		Rosenblum~\cite{Rosenblum11} & 2011 & x86-64 & No & CRF, SVM & --- & Yes & Yes & 3 & 3 & 2 & Function \\
		Rahimian~\cite{Rahimian15} & 2015 & x86-64 & No & Uncertain$\mathparagraph$ & --- & Yes & No & 4 & 2 & 2 & Binary \\
		Chen~\cite{Chen18RNNOpti} & 2018 & x86-64 & Yes & LSTM & No & No & No & --- & --- & 4 & Function \\
		Yang~\cite{Yang19CNNOpti} & 2019 & ARM32 & Yes & CNN & No & No & No & --- & --- & 4 & Object File \\
		Massarelli~\cite{Massarelli19} & 2019 & x86-64 & Yes & GNN, RNN & Yes & Yes & No & 3 & --- & --- & Function \\
		Otsubo~\cite{Otsubo20} & 2020 & x86-64, x86 & Yes & CNN & Yes & No & No & 4 & 1-2 & 2 & Binary \\
		Pizzolotto~\cite{Pizzolotto20} & 2020 & x86-64 & Yes & CNN, LSTM & No & No & No & 2 & --- & 2 & Chunk \\
		Benoit~\cite{Benoit21} & 2021 & x86-64 & Yes & GNN & No & Yes & No & 4 & 5-6 & 4 & Binary \\
		Ji~\cite{Ji21} & 2021 & x86-64 & Yes & GNN & Yes & Yes & Yes & 2 & 3 & 4 & Binary \\
		Tian~\cite{Tian21} & 2021 & x86-64 & Yes & CNN, RNN & Yes & No & Yes & 3 & 1-4 & 3 & Function \\
		Pizzolotto~\cite{Pizzolotto21} & 2021 & Agnostic & Yes & CNN, LSTM & No & No & No & 2 & --- & 5 & Chunk \\
		DIComP~\cite{dicomp} & 2022 & x86-64 & Yes & MLP & No & No & Yes & 2 & 2-3 & 5 & Binary \\
		Otsubo~\cite{Otsubo22} & 2022 & Agnostic & Yes & CNN & Yes & No & No & 2-4 & 1-2 & 2 & Chunk \\ 
		Du~\cite{Du22} & 2022 & x86-64 & No & Boosting & No & No & Yes & 3 & 1-2 & 3 & Function \\
		He~\cite{BinProv} & 2022 & Agnostic & Yes & Transformer & Yes & No & No & 2 & --- & 2-4 & Chunk \\
		\hline
		Our study & 2023 & ARM32/64 & No & Linear SVM & --- & No & Yes & 3 & 1-2 & 2-5 & Binary \\
		\hline
	\end{tabular}
	\vspace*{1em}
	
	$*$ NN? indicates if the model is a neural network. $\dagger$ The presence or absence of the attention technique applies to neural networks only. $\ddagger$ This column indicates if register use is considered for feature extraction. $\mathsection$ These columns indicate the size of the label space for compiler family, version, and optimization levels. $\mathparagraph$ The classifier is not specified explicitly in this work, but is reported as hierarchical.
	\vspace{-1.5em}
\end{table*}

\section{Background and Related Work} \label{s:related-works}
We discuss and compare the contributions of preceding works, especially DIComP~\cite{dicomp} and Pizzolotto et al.~\cite{Pizzolotto21}.

\parhead{Compiler Provenance as a Classification Problem.}
Compiler provenance recovery is a budding application of machine learning models to stripped binary analysis.
The seminal work of Rosenblum et al.~\cite{Rosenblum10} predicts the compiler
family for x86-64 binaries using a conditional random field (CRF). Their second
work~\cite{Rosenblum11} follows with control flow graph (CFG)-derived features, adding a SVM to recover the \textit{full} compiler
provenance (family, version, and optimization).
Rahimian et al.~\cite{Rahimian15} use a hierarchical classifier, where the compiler family prediction affects the inference for the other labels.

\parhead{Applications of Deep Learning.}
Subsequently, deep learning models were applied, minimizing the models' dependency on handcrafted features.
Instead, these models attempt to learn features from instruction-to-matrix embeddings.
Chen et al.~\cite{Chen18RNNOpti} and Yang et
al.~\cite{Yang19CNNOpti} use recurrent (RNN) and convolutional (CNN) neural networks for this, respectively.
Notably, Yang et al. are the
first to address the ARM architecture (with all prior
works targeting x86 or x86-64).
Tian et al.~\cite{Tian21} follow with a hybrid of RNNs and CNNs, and He et al.~\cite{BinProv} use transformers, an improved version of RNNs now commonly used for natural language processing.

Furthermore, graph neural networks (GNNs)~\cite{Scarselli08GNN, Micheli09GNN}
were applied to learn features from CFGs of binaries.
Massarelli et al.~\cite{Massarelli19} use a GNN to recover the compiler
family, and also applies attention~\cite{Vaswani17Attention}, a
technique to emphasize learning on a subset of the input.
Benoit et al.~\cite{Benoit21} follow with a hierarchical GNN model that recovers the full compiler provenance.
Ji et al.~\cite{Ji21}
augment CFG- and instruction-derived features, embedding them as input to a GNN.

\parhead{Diversity in Recent Approaches.}
More recently, diverse inference techniques have been applied to compiler provenance recovery. 
In contrast to prior works performing static analyses, Du et al.~\cite{Du22} execute individual functions in isolation with a binary emulation library to execution traces as features, then classify them with a gradient-boosting algorithm.
Moreover, some recent models by Otsubo et al.~\cite{Otsubo20, Otsubo22} and Pizzolotto et
al.~\cite{Pizzolotto20} aim to recover the full compiler provenance from a
short sequence of information from a binary.
Notably, \cite{BinProv, Otsubo22, Pizzolotto20} embed raw bytes from a binary's executable section without
any consideration of architecture.

\parhead{Replicating ARM-based Models.}
To our knowledge, three works have addressed compiler provenance on ARM binaries.
Yang et al.~\cite{Yang19CNNOpti} focus mainly on feature engineering for ARM assembly than developing classifiers for different labels.
While the work of He at al.~\cite{BinProv} is the most recent, we were unable to find code or data artifacts for replication.
On the other hand, the follow-up work of Pizzolotto et al.~\cite{Pizzolotto21} contributes an extended evaluation of their architecture-agnostic model on binaries from various ISAs, including ARM.
We use this work as one baseline to compare the performance of DIComP's feature engineering approach against a deep-learning model.

\parhead{Our Baseline for a Deep-learning Model.}
The feature extraction technique of the Pizzolotto model is one the authors describe as
``naive''~\cite[Section III-B]{Pizzolotto20}, but is demonstrably
accurate ($>$90\%) for multiple architectures. The technique dumps the executable
section of each binary, and splits it into fixed-size chunks that range from 1
to 2048 bytes (our comparison in \cref{s:evaluation} is against the full 2048 bytes).

The classifier is either a CNN or a long short-term memory (LSTM) model, a type of RNN~\cite{Hochreiter97LSTM}. While the LSTM outperforms the CNN by
a small gap (within 8\%) in accuracy, its end-to-end runtime is up to 6 times
slower. Therefore, our evaluation uses the CNN, given its advantage in speed
for a less notable sacrifice in accuracy. The Pizzolotto model evaluates both
models on binaries from 7 architectures, including 32- and 64-bit ARM. We base
our evaluation on their corpus of 64-bit ARM binaries, since it offers the
broadest label space of 2 compiler families and 5 optimization levels. 

\parhead{DIComP: Our Baseline for a Lightweight Model.}
In contrast to a series of works using deep learning, DIComP~\cite{dicomp} is a recent model demonstrating that a handcrafted feature set and lightweight classifier can outperform prior models in not only accuracy, but most notably speed.
By disassembling ground-truth x86-64 binaries from identical source code but varying in compiler family, version, or optimization level, the authors observe that the compiler provenance is distinguishable through unique patterns in the opcodes (also called mnemonics) and register usage of the binaries' instructions.

Accordingly, DIComP derives features from opcodes, registers, and function boundaries to train a hierarchy of multi-layer perceptrons (MLP).
While they are a type of neural network, MLPs are among the lightest, with a fraction of the complexity of typical CNNs, RNNs, and GNNs.
Despite the simple structure, DIComP achieves perfect accuracy on predicting the compiler family, and high accuracy for optimization level.

Hence, our replication study focuses on the feature engineering of DIComP, investigating if opcode- and register-derived features are also effective for ARM binaries. 
We take more steps towards a lightweight model by replacing the MLPs with linear SVMs and forgoing the function boundary feature (which requires a reverse engineering tool).
We compare this revised model on the dataset and model code of our deep-learning baseline from Pizzolotto et al.~\cite{Pizzolotto21}

\parhead{State of the Art in Compiler Provenance.}
Finally, we show a summary of all previous works in compiler provenance
recovery we have discussed, and contrast them by a number of key metrics
in~\cref{f:SoK}.

\begin{figure}[htb]
	\vspace{-0.5em}
	\includegraphics[width=\linewidth]{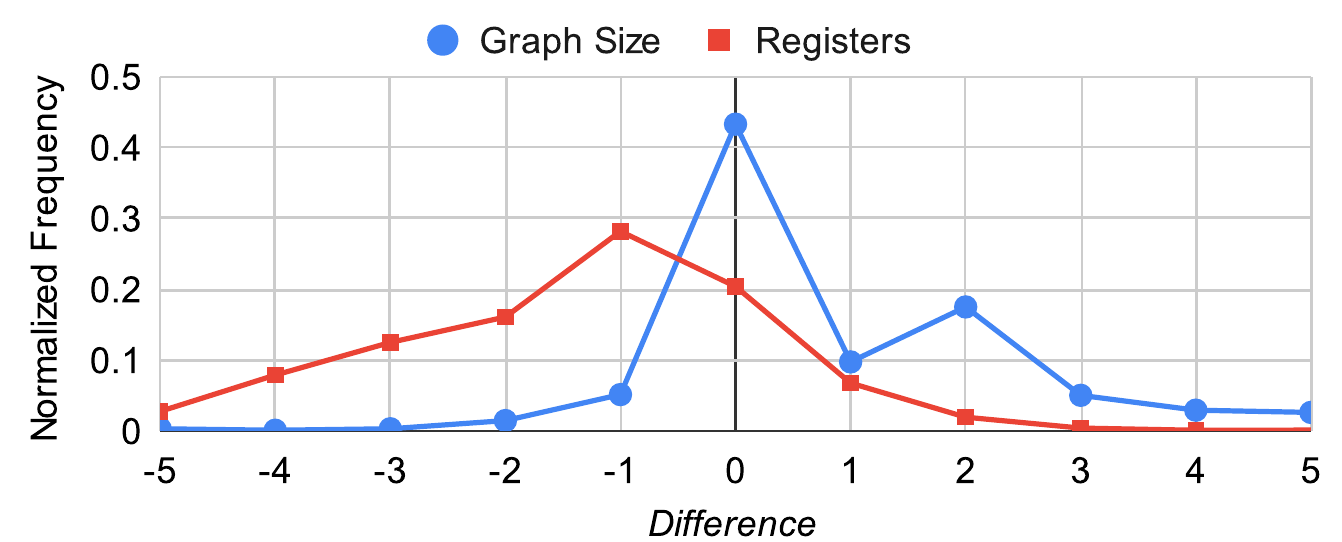}
	\caption{Frequency distribution of the \emph{Difference} in CFG size and \# registers used in each binary's \texttt{main} function.}
	\label{f:stats}
	\vspace{-1em}
\end{figure}

\begin{figure*}[htb]
	\centering
	\includegraphics[trim={0cm 0cm 0cm 0cm}, width=0.8\linewidth]{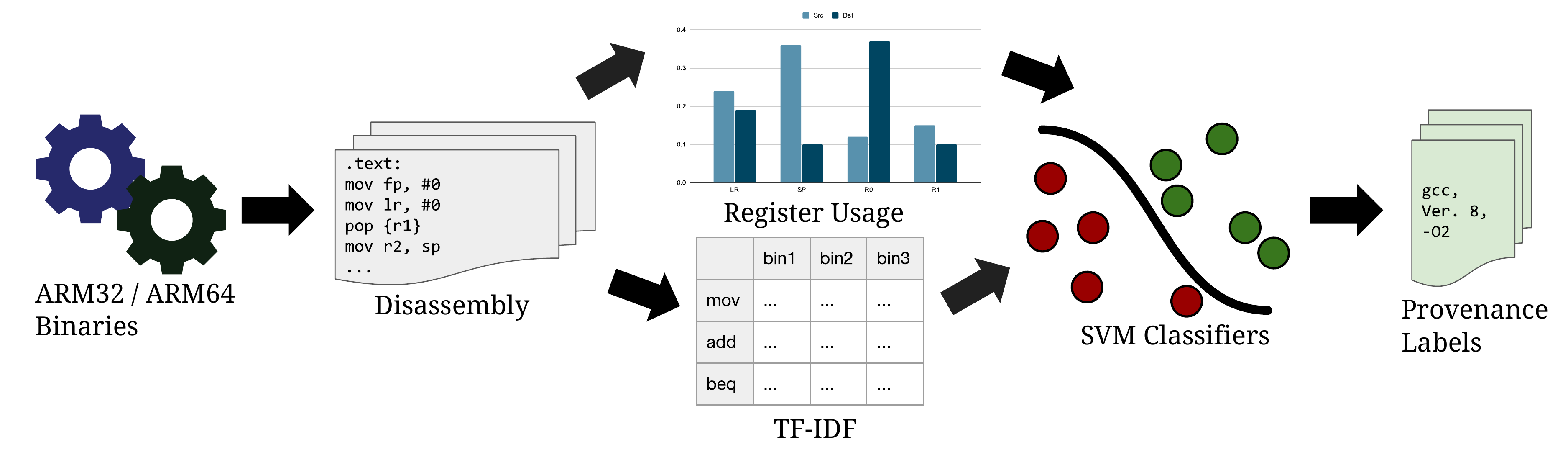}
	\vspace{-0.5em}
	\caption{Overview of our replication of DIComP~\cite{dicomp} with minor revisions to the methodology. We first disassemble the binaries.
	Then, we generate frequency distributions for register usage, and score opcodes by TF-IDF to emphasize patterns in less common opcodes.
 	We concatenate the two distributions and the score vector for classification.}
	\label{f:model-overview}
	\vspace{-1em}
\end{figure*}

\section{Motivation}\label{s:comparison}
We now discuss our motivation for reproducing DIComP's opcode- and register-focused features on ARM binaries instead of x86-64 binaries.
Noticing from \cref{f:SoK} that several x86-64 models performed CFG extraction to derive features, we aim to determine how suitable graph-based and instruction-based analyses are for a fast model working on ARM binaries.

To do so, we compile pairs of x86-64 and ARM binaries from the same source code,
then profile them based on their CFG and instructions separately. Next, we discuss the differences between the binaries which manifest in each metric, and also whether that type of analysis is apt for ARM binaries.

\parhead{Experimental Setup.}
We collect 30 open-source projects from GitHub with C or C++ source code,
and compile the projects with identical build configurations on x86-64 and
32-bit ARM machines. We use 32-bit ARM for this comparison due to the ISA
having a similar number of general-purpose registers as
x86-64~\cite{ARMRegisters, x64Registers}. We compile the x86-64 binaries on a system with an
Intel i5-8250U CPU, 8 GB of RAM, and Ubuntu 21.04, and the ARM binaries on a
Raspberry Pi 4 Model B with 2 GB of RAM.

We compile each project several times across several compiler types and options,
permuting the label space of the binaries on each pass. We use gcc versions 6
and 8, Clang versions 7 and 9, and optimization levels O0 and O2. After
compilation, we remove binaries with duplicate checksums, where most are
`toy' binaries produced by the build system (e.g., CMake and \texttt{configure}
scripts). This results in 976 pairs of x86-64 and 32-bit ARM binaries. For
profiling, we consider only the \texttt{main} function of each binary to
reduce discrepancies between large and small binaries, and for experimental
expediency.

We use angr~\cite{Wang17Angr} to extract the \texttt{main} function's CFG and profile the number of basic blocks in it (i.e., graph size), as an example of a CFG-derived metric.
Also, we use radare2~\cite{Radare2} to count the number of distinct registers to exemplify an instruction-based metric.
We define a unit, \emph{Difference}, as the
metric in the x86-64 binary in each pair minus that in the 32-bit
ARM binary, i.e., $D = x_{\texttt{x86-64}} - x_{\texttt{ARM}}$. For each value
of $D$ on the x-axis, we report the normalized frequency of pairs of binaries
exhibiting that difference on the y-axis. For example, the coordinate $(2,
0.18)$ on the blue line indicates that the graph size is greater by 2 in the
x86-64 binary than the ARM binary in 18\% of the pairs.

\parhead{Results.}
We plot the frequency distributions in \cref{f:stats}. In the blue plot for
graph size, the ARM binary has
1-3 fewer basic blocks in more than 30\% of pairs. This suggests that the
x86-64 ISA may allow a routine to run conditionally in some cases,
whereas it must run linearly on ARM.

Furthermore, in the red plot for the
number of distinct registers used, the ARM binary uses 1-5 more registers in
more than 70\% of pairs. This indicates that x86-64 is less
likely to use the full set of registers, instead spilling and reusing
registers in a smaller working set. Since registers \texttt{r8} through
\texttt{r15} in x86-64 have no counterparts in the 32-bit x86 architecture, we
hypothesize the reason may be legacy or compatibility reasons in compiler
backends. Furthermore, with 64-bit ARM having roughly double the general-purpose registers~\cite{ARM64Registers} of 32-bit ARM, we expect an even larger working set and thus greater variety.

\parhead{Insights for Classifying ARM Binaries.}
We find angr's runtime to be prohibitive for a fast model, taking more than 10 minutes for some binaries.
In contrast, radare2 completes in a few seconds at most to count registers, motivating our replication study towards instruction-based analysis akin to DIComP~\cite{dicomp}.
Furthermore, given the smaller CFG size and greater register coverage of ARM binaries, we hypothesize DIComP's technique for x86-64 binaries will transfer well.

\section{A Lightweight Feature Set for ARM Binaries} \label{s:technical}  
In this section, we present our approach to predict the compiler provenance of
a stripped ARM binary.
Following the approach of DIComP~\cite{dicomp},
we first preprocess input binaries into their
disassembly, a much faster task than graph extraction which proved to be prohibitive (cf. \cref{s:comparison}).
Also noticing that the ARM ISA provides more freedom in register selection from \cref{s:comparison}, we make frequency distributions over all registers when each is used as a source or destination.

Subsequently, we make minor adaptations to DIComP's approach.
Instead of picking a fixed set of opcodes per classification task~\cite{dicomp}, we adapt the TF-IDF metric to automatically score opcodes that stand out.
Moreover, instead of a hierarchy of MLP classifiers, we replace the MLPs with linear SVMs for a more lightweight model and greater transparency of the decision boundary.
To that aim, we do not follow DIComP's function boundary feature, as it uses an LSTM model, adding significant bulk to the model's weight.
Similarly to DIComP, the SVMs predict the compiler family, version, and optimization
level. See \cref{f:model-overview}.

\parhead{Preprocessing.}
Following DIComP~\cite{dicomp}, we use the \texttt{objdump} utility disassemble all executable sections of the input binary.

\parhead{Profiling Register Usage.}
Subsequently, we collect the relative frequency distributions of source operands and destination operands over all general-purpose registers.
For instance, the two operands and one destination for an \texttt{add} instruction have different purposes, and thus we profile them separately.

For 64-bit ARM assembly, registers can be referenced by all 64 bits, or their lower 32 bits.
We treat them as separate cases since their usage by the compiler may differ. For example, a register may hold a pointer in the former in contrast to an \texttt{int} in the latter.
Furthermore, we use the ratio of references of \texttt{fp} to references of \texttt{sp} as an additional feature, after observing that compilers use either to reference local variables.

\begin{figure}[t]
	\centering
	\vspace{-0.5em}
	\includegraphics[trim={0cm 0cm 0cm 0cm}, width=0.9\linewidth]{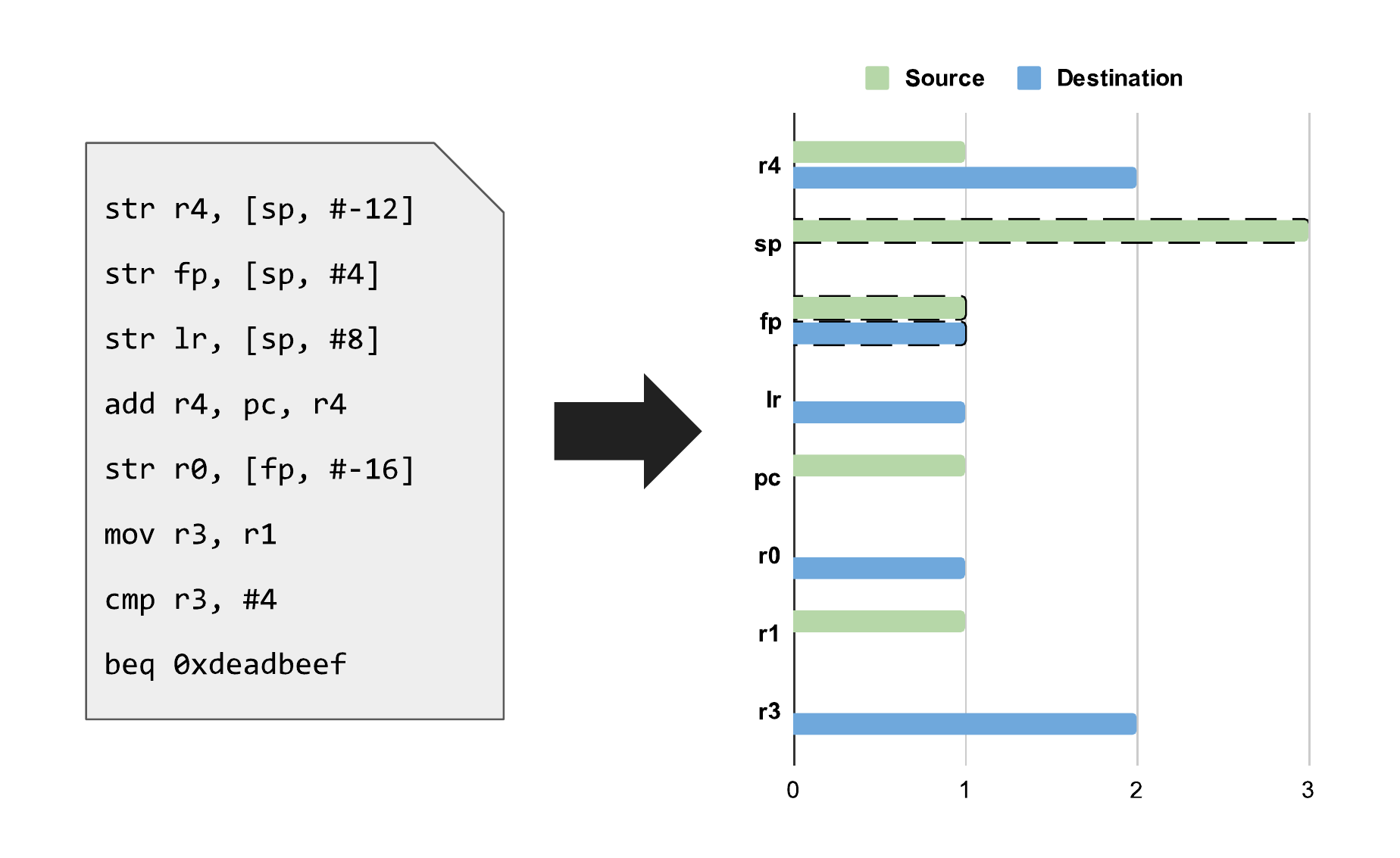}
	\vspace{-1.5em}
	\caption{Motivating example and its source and destination register frequency distributions. The frequencies for \texttt{fp} and \texttt{sp} are shown as bars with dashed lines. Unused registers are omitted for brevity but have frequency zero.}
	\label{f:register-usage}
	\vspace{-1.5em}
\end{figure}

Finally, we show a motivating example of how a snippet of ARM assembly (using 32-bit ARM for simplicity) is represented in the feature space in \cref{f:register-usage}.

\parhead{Profiling Opcodes with TF-IDF.}
\label{s:profiling-opcodes}
We now focus on the opcodes from the preprocessing step.
DIComP~\cite{dicomp} uses a fixed set for each classification task, which they determine by analyzing binaries of their training corpus with reverse engineering tools.
In contrast, we use TF-IDF (term frequency-inverse document frequency), a
statistic from information retrieval to quantify the relative importance of
each word in a document given a collection of documents. TF-IDF is commonly
used by search engines to determine the most relevant results to a query, which we adapt instead of manually picking opcodes that are distinguishable across compiler provenances.

We view the binary's opcode sequence as a document of words, and
repurpose TF-IDF such that `words' are opcodes and `documents' are ARM binaries
in our corpus.  We hypothesize that rare opcode usage patterns in a binary
would result in statistics that are distinct enough to be separated by our
classifier. Given that TF-IDF puts emphasis on words that occur frequently in a
document yet seldom occur in the collection of documents, we attempt to
identify the opcodes distinctly preferred by each compilation environment. A
common example in ARM would be zeroing a register \texttt{reg}, which a compiler backend can emit as either \texttt{mov reg, \#0} or \texttt{eor reg, reg}, among many others.

For each binary in the corpus, we calculate the relative term frequency for all
opcodes present in the disassembly. That is, for opcode $i$ in a binary with
$k$ instructions, $tf_i = \frac{count(i)}{k}$. Next, we count the occurrences
of each opcode in the entire corpus to calculate the inverse document
frequency. If $N$ is the size of the corpus and $n_i$ is the number of binaries
in which opcode $i$ appears, then $idf_i = \log \frac{N}{n_i}$.

\begin{figure}[htb]
	\vspace{-1em}
	\begin{center}
		\includegraphics[trim={0cm 0cm 0cm 0cm}, width=0.9\linewidth]{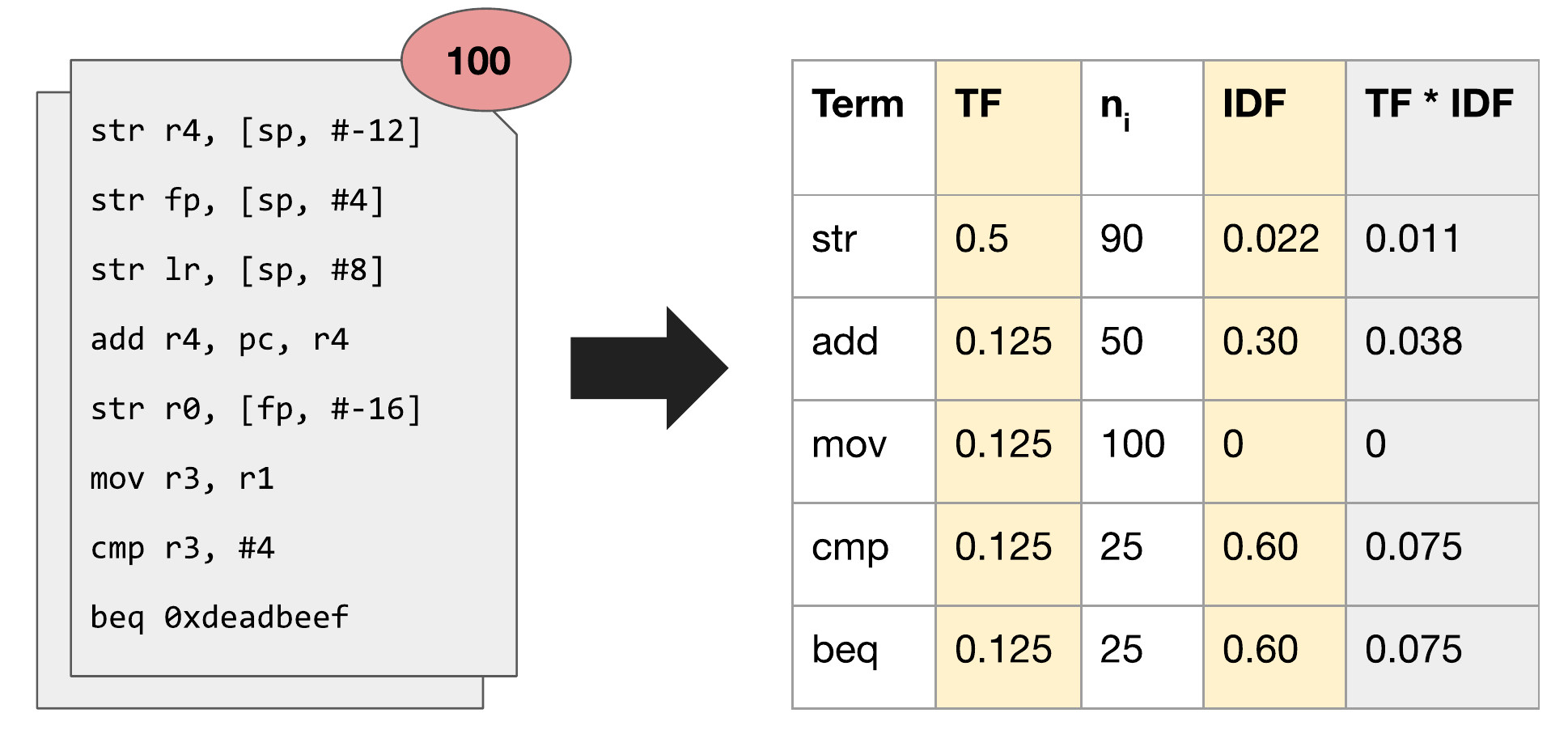}
	\end{center}
	\vspace{-0.5em}
	\caption{Example TF-IDF calculation. For simplicity, we assume the number of binaries containing each opcode shown in the table above, and a corpus size of 100 binaries including the ARM assembly snippet. The final $tf*idf$ column (in grey) is the product of the two columns highlighted in yellow.}
	\label{f:tf-idf}
\end{figure}

Multiplying $tf_i$ over all opcodes $i$ present in the corpus with its corresponding
$idf_i$ results in a fixed-size vector of TF-IDF scores for each binary. If an
opcode is not present in a binary at all, we treat its term frequency as zero.
We then take the L2, or Euclidean, norm of the vector before forwarding it to
the classifier. As an example, we show these calculations used for a snippet of
32-bit ARM assembly in \cref{f:tf-idf}.

\begin{figure}[htb]
	\centering
	\includegraphics[trim={0cm 0cm 0cm 0cm}, width=0.9\linewidth]{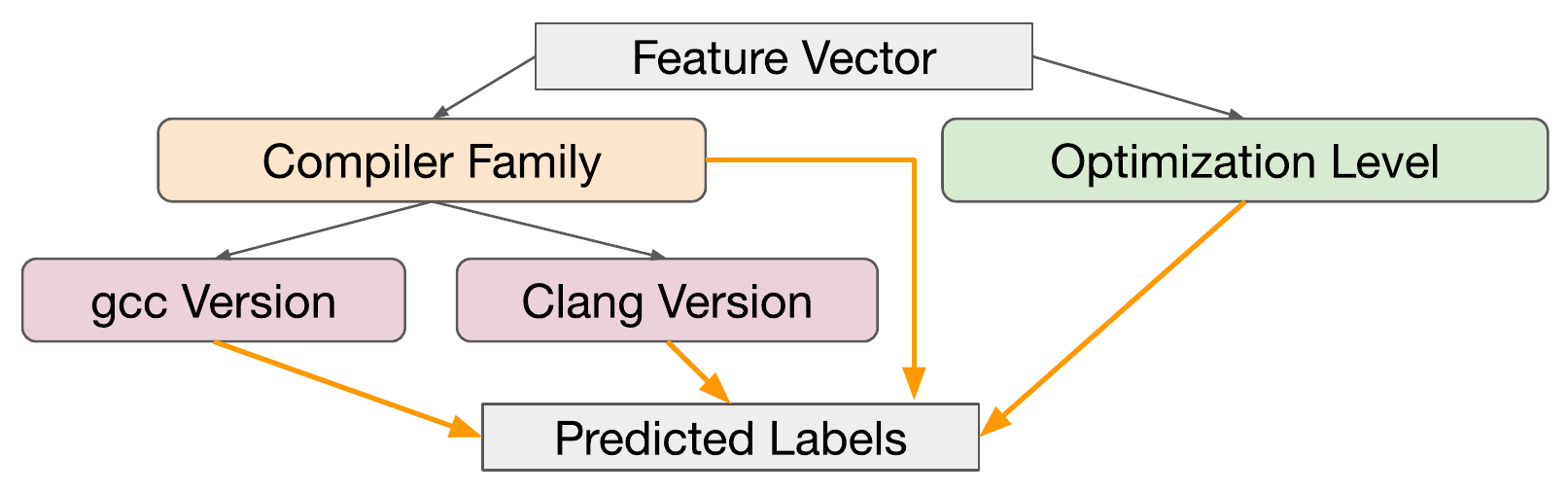}
	\vspace{-0.5em}
	\caption{Organization of SVMs in our model. Gray arrows indicate the feature vector being relayed, while the orange arrows indicate resulting predictions.}
	\label{f:hierarchy}
	\vspace{-1em}
\end{figure}

\parhead{Classification.}
We concatenate the two frequency distributions of register usage with the TF-IDF scores of each binary.
We then classify the features with a hierarchy of four linear-kernel SVMs using L1 penalty. See \cref{f:hierarchy}.
This follows the classifier of DIComP~\cite{dicomp}, but replaces the MLP (a simple neural network) with an even more lightweight SVM.

We run 10-fold cross validation and set their class weights inversely proportional to the class frequency to combat overfitting.
In each fold, we first classify the compiler family and optimization level in parallel.
Following DIComP's hierarchy, the family prediction determines if the features should be handled by a version classifier for gcc or for Clang.
That is, we expect the features for each compiler family to differ notably, since a subtle difference in which registers a program favors to use or which opcode the compiler picks will translate to different features in aggregate.

Our model outputs a tuple of the predicted compiler family, version, and optimization labels.
We apply this model to several experimental corpora of ARM binaries to measure its accuracy and speed against our deep-learning baseline model of Pizzolotto et al.~\cite{Pizzolotto21}, and then to large repositories of ARM binaries to measure scalability.

\section{Evaluation} \label{s:evaluation}
We recall that our goal is to recover the compiler provenance of unknown and stripped ARM binaries with a model derived from DIComP~\cite{dicomp} that is accurate, fast, and interpretable.
In this section, we gauge the performance of our model explained in \cref{s:technical} along these three criteria for 32- and 64-bit ARM binaries, to determine if DIComP's approach is reproducible.

We use a corpus of 64-bit ARM binaries from the work of Pizzolotto et al.~\cite{Pizzolotto21} and the model source code for that work to perform a direct comparison against our model.
We show that our lightweight approach achieves accuracy on par with the state-of-the-art CNN-based Pizzolotto model, while training 583 times and classifying samples 3,826 times faster.

In addition, by inspecting weight vectors of each SVM component in our model, we develop insights about the most salient features that distinguish compiler families and optimization levels, mirroring DIComP's analysis using reverse engineering tools. These are observations which are difficult to deduce in state-of-the-art neural networks, since they apply nonlinear transformations on the previous state at each layer.

Subsequently, we show that our model remains accurate and fast on our own corpus of 32-bit ARM binaries compiled over a larger label space, including binaries from a formally-verified compiler and binaries from multiple compiler versions within each family.
Once more, we identify the features which help our model classify the most accurately, and discuss how they differ from their counterparts on 64-bit ARM.

Finally, we show that the DIComP-based lightweight technique is scalable to large binary corpora
by predicting the compiler provenance of binaries from the package repositories and installation images of three widely-used Linux distributions for ARM.
We performed all experiments on a ThinkPad X1 laptop with an Intel i7-10710U CPU, 16 GB of RAM, and Ubuntu 21.04, unless mentioned otherwise.

\subsection{Evaluation on 64-bit ARM}
\label{s:benchmark}
We compare our model's performance against our deep-learning baseline model~\cite{Pizzolotto21}, then show how it achieves the same accuracy while executing faster and allowing developers to better understand its decision-making.
We reuse the publicly-available dataset from Pizzolotto et al.~\cite{Pizzolotto21}, 
which contains 15,925 samples across a label space of \{gcc,
Clang\} for the compiler family and \{-O0, -O1, -O2, -O3, -Os\} for the
optimization level.  We use the CNN model presented in Pizzolotto et al.~\cite{Pizzolotto21} for comparison as it is
much faster with a minimal loss in accuracy. On the CNN, their paper claims an
accuracy of 0.9996 for the compiler family and 0.9181 for the optimization
level~\cite[Section V-A]{Pizzolotto21}.  
We reproduced the results using their pretrained CNN as-is to achieve accuracies of 0.9988 and 0.9167 respectively.

\begin{figure}[t]
	\centering
	\vspace{-1em}
	\includegraphics[trim={0cm 0cm 0cm 0cm}, width=\linewidth]{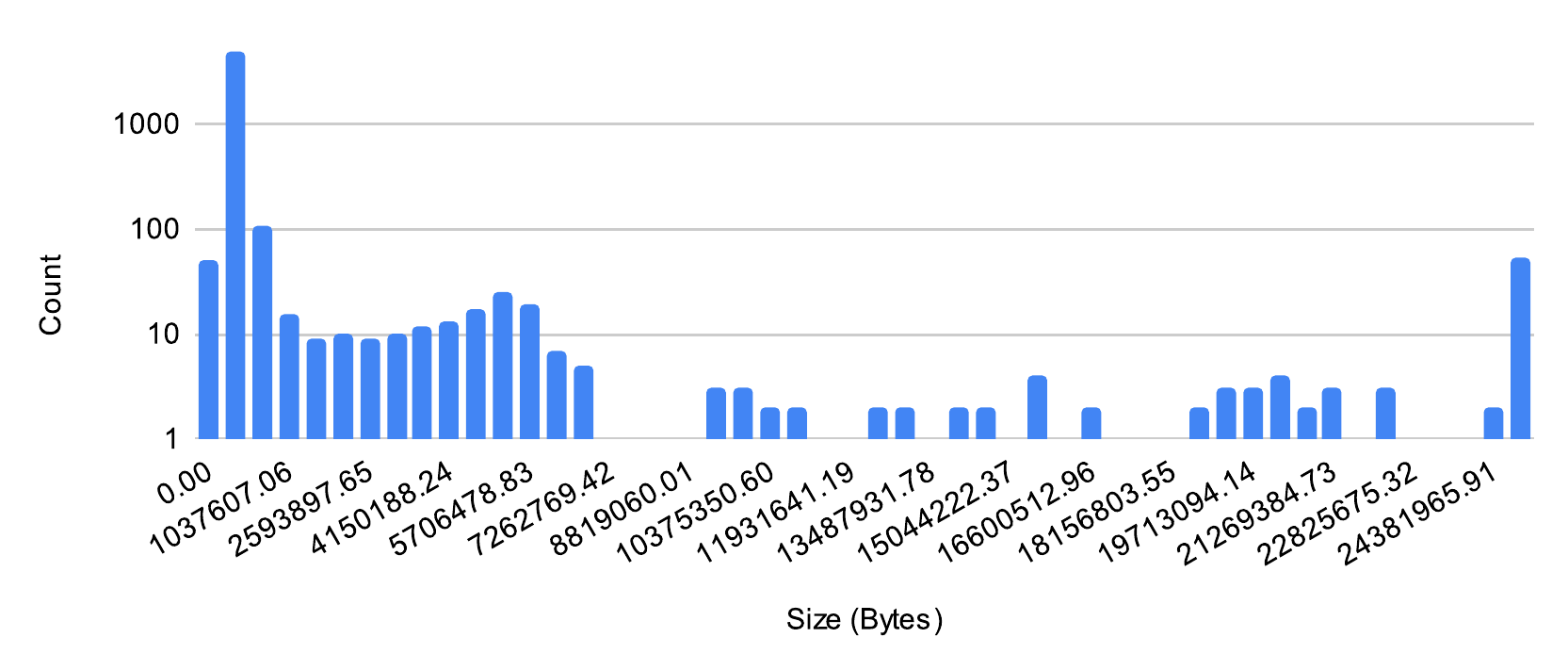}
	\vspace{-2em}
	\caption{Histogram of binary sizes in the original 64-bit ARM dataset from the work of Pizzolotto et al.~\cite{Pizzolotto21}}
	\vspace{-1.5em}
	\label{f:before-filtering}
\end{figure}

\parhead{Binary Size Imbalance Leads to Overfitting.}
We recall from \cref{s:related-works}  that the preprocessing step of the
Pizzolotto model consists of taking the executable section of an input binary
and then dividing it into 2048-byte chunks. Each chunk then becomes an input to
the CNN. A consequence of this step is that the number of samples provided to
the CNN by each binary is proportional to its size.

During training, if the training set contains a disproportionately large binary,
the model risks being overfit to classify the large binary instead
of developing a generalizable decision binary that applies to other smaller
binaries. This is because correctly classifying 2048-byte chunks from the
large binary minimizes the training loss. In contrast, we take one sample per binary to avoid overfitting. We 
show the distribution of binary sizes from the 64-bit ARM dataset in \cref{f:before-filtering}.

The histogram indicates that most binaries are smaller than 1 MB in size.
However, it also shows a long tail of binaries exceeding it (note the y-axis is logarithmic), many of them disproportionately large in size.
In raw numbers, the 50 largest binaries have a combined size of 2.1 GB.
Conversely, the remaining 15,875 binaries have a combined size of 1.9 GB.
That is, the 50 largest binaries contribute more data to the model than all other binaries.
We demonstrate this imbalance leads to overfitting below, and rectify it in our benchmark.

\begin{figure}[t]
	\centering
	\vspace{-1em}
	\includegraphics[trim={0cm 0cm 0cm 0cm}, width=\linewidth]{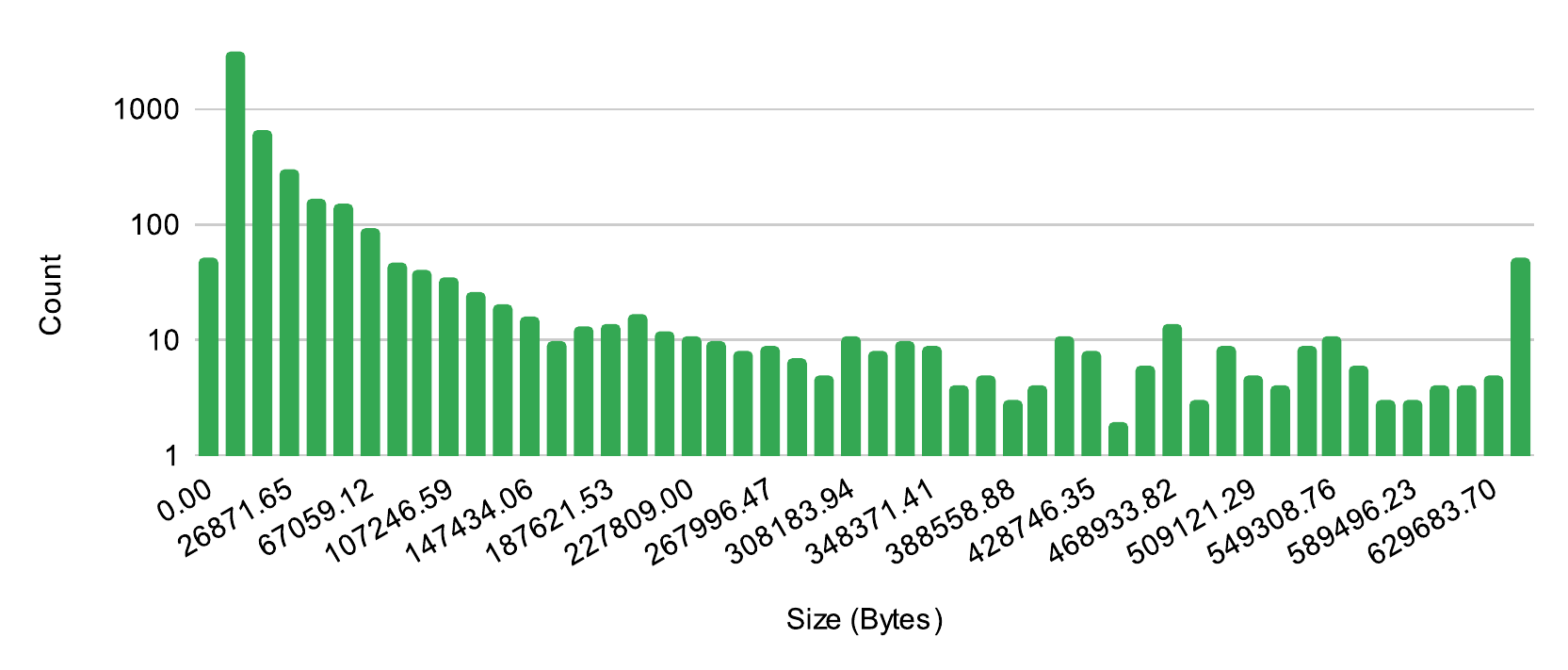}
	\vspace{-2em}
	\caption{Histogram of binary sizes after filtering binaries larger than 1 MB.}
	\vspace{-1em}
	\label{f:after-filtering}
\end{figure}

\parhead{Experimental Setup With Dataset Filtering.}
Since the majority of binaries are smaller than 1 MB in this data set, we removed binaries larger than 1 MB from consideration.
We also removed binaries with duplicate checksums (cf. \cref{s:comparison}) from this dataset, ultimately reducing it to 14,140 binaries smaller than 1 MB.
We show a histogram of binary sizes on this filtered dataset in \cref{f:after-filtering}.

We note that, compared to \cref{f:before-filtering}, the inequality between large and smaller binaries is much less pronounced.
We use this filtered dataset to evaluate the CNN model and our replication of DIComP on two metrics: accuracy and end-to-end runtime for training followed by evaluation (from preprocessing the binaries to outputting predictions).

We use the Linux \texttt{time} utility to measure the latter, where we take the sum of the \texttt{user} and \texttt{sys} times (all units are seconds).
We note this is not wall clock time.

\parhead{Accuracy and Runtime Results.}
First, we use the pretrained model on the filtered dataset to determine if the large binaries had overfitted the model.
The compiler family accuracy decreases sharply from 0.9988 to 0.0212. Although less striking, the optimization level accuracy decreases substantially from 0.9167 to 0.7290.
Next, we use Pizzolotto model code as-is without modifications to train the CNN on our corpus of binaries less than 1 MB in size.

On evaluation, the retrained CNN achieves accuracies of 0.9852 on the compiler family and 0.7515 on the optimization level, comparable to the numbers reported on the work of Pizzolotto et al.~\cite[Section V-A]{Pizzolotto21} for the former, but a substantial decrease for the latter.
\emph{We regard these accuracy figures as baselines for accuracy against our model}.

With our model, using the same number of classes for compiler family (2: gcc and Clang) and optimization level (5: Os, O0, etc.), we observed mean accuracies of 0.9991 and 0.7180 respectively on 10-fold cross validation, putting it on par with the CNN model with regard to accuracy.
In comparison to DIComP~\cite{dicomp}'s results, we almost match their perfect accuracy on distinguishing gcc and Clang.
DIComP evaluates its optimization level classification performance separately per compiler on precision, recall, and F1-score, with all three metrics at about 0.90 for gcc and 0.70 for Clang.
While this makes a direct comparison more nebulous, our model shows their technique generally transfers well to ARM binaries without significant drops in performance.

Like DIComP, the lightweight structure of our model contributes substantially to the performance of compiler provenance models with regard to runtime.
As shown in \cref{t:pizzolotto-comparison}, we observe a 583-times speedup in training time on each binary, and a 3,826-times speedup on evaluating it. The only portion of the end-to-end benchmark where we did not observe orders of magnitude improvements 
is feature extraction. Though, we note this step is bound by disk read speeds, since either model must load the binary into memory.

These speedups also lead to pronounced reductions in energy use, a problem that deep neural models continue to face (cf. \cref{s:intro}).
Overall, we show with this benchmark that DIComP's technique~\cite{dicomp} is reproducible on ARM binaries.

\begin{table}[htb]
	\centering
	\def\arraystretch{1.5}
	\vspace{-0.5em}
	\caption{Comparison of accuracy and end-to-end runtime between our DIComP-based model~\cite{dicomp} and the CNN model of Pizzolotto et al.~\cite{Pizzolotto21} \label{t:pizzolotto-comparison}}
	\footnotesize
	\begin{tabular}{l|rrr}
		\hline
		\textbf{Metric} & \textbf{Our Model} & \textbf{CNN~\cite{Pizzolotto21}} & \textbf{Advantage} \\
		\hline
		Compiler Accuracy & 0.9991 & 0.9852 & 0.0139 \\
		Optimization Accuracy & 0.7180 & 0.7515 & -0.0335 \\
		\hline
		Training per Binary & 15.80 ms & 9.21 s & 583.13x \\
		Extraction per Binary$*$ & 130.15 ms & 225.64 ms & 1.73x \\
		Prediction per Binary & 1.15 $\mu$s & 4.40 ms & 3,826.09x \\
		\hline
	\end{tabular}
	\vspace{1em}
	\\
	$*$This step entails reading the binaries from disk and thus is I/O-bound.
	\vspace{-0.5em}
\end{table}

\parhead{Experimental Setup for Model Interpretation.}
Since we use a hierarchy of linear-kernel SVMs, the model weights for each feature translate directly to the coefficients of the hyperplane decision boundary, giving us insights of which features are the most important for the predictions.
In contrast, neural networks are often notoriously difficult to interpret in this manner due to the sheer volume of layers, parameters, and nonlinear transformations (e.g., Sigmoid or ReLU).

After training each SVM in the hierarchy (recall \cref{f:hierarchy}), we use the ELI5 library~\cite{ELI5} to map feature names to dimensions in the feature vector on the first cross-validation fold, and then obtain the five features with the largest learned weights (by magnitude).
Mirroring DIComP~\cite{dicomp}, we apply this methodology throughout this section to develop insights about how compiler provenance impacts code generation.

\parhead{Insights from Model Interpretation.}
The dataset from this subsection specifies a binary classification task for the compiler family, but a 5-way task for the optimization level.
In the latter task, we use five one vs. rest classifiers.
Hence, we show the five heaviest weights by magnitude and their corresponding feature names for the six linear-kernel SVMs in \cref{t:pizzolotto-interp}
(the greater the magnitude, the greater the influence on prediction).
For the feature names, the prefix \texttt{src\_} indicates the feature is a register being used as a source operand, and \texttt{dst\_} as a destination operand.
For the registers, 64-bit register names start with \texttt{x}, while their 32-bit counterparts (i.e, lower 32 bits of each register) start with \texttt{w}.

When classifying the compiler family, we note that both \texttt{x8} and \texttt{w8} are heavily weighed. They are the indirect result register~\cite{ARM64Registers} used to pass pointers between subroutines.
Furthermore, \texttt{wzr} is a `zero' register, where reads are 0 and writes are discarded.
Accordingly, the model may distinguish Clang from gcc by where each compiler stores function parameters or return values, or by the compiler's use of zero registers.

For the optimization level, we observe \texttt{nops} are important for every label except \texttt{-O3}, in addition to the presence of both frame and stack pointers (\texttt{x29} and \texttt{sp} respectively) and the \texttt{sdiv} (signed divide) instruction.
This can be explained by \texttt{nops} being rare in higher optimization levels (especially \texttt{-Os} to reduce binary size), and frame and stack pointers being used less commonly when small functions become inlined.

\begin{table}[htb]
	\centering
	\def\arraystretch{1.5}
	\vspace{-0.5em}
	\caption{Heaviest weights of each SVM and their corresponding features when trained on the filtered 64-bit ARM dataset. Register features are italicized. \label{t:pizzolotto-interp}}
	\footnotesize
	\begin{tabular}{ll|rrrrr}
		\hline
		\textbf{Task} & & \textbf{1} & \textbf{2} & \textbf{3} & \textbf{4} & \textbf{5} \\
		\hline
		Family & Weight & 28.347 & 20.164 & 17.607 & 4.249 & 3.183 \\
		& Feature & \textit{dst\_x8} & \textit{src\_wzr} & \textit{dst\_w8} & stur & b.lt \\
		\hline
		-O0	& Weight & 12.131 & 5.085 & -4.977 & -3.996 & 3.732 \\
		& Feature & nop & dmb & cbz & \textit{src\_x19} & \textit{src\_sp} \\
		-O1 & Weight & -35.996 & -27.139 & -19.041 & 18.799 & -15.941 \\
		& Feature & nop & sdiv & \textit{dst\_sp} & \textit{src\_x30} & \textit{src\_w8} \\
		-O2 & Weight & 20.589 & -13.904 & 12.474 & -7.218 & 6.719 \\
		& Feature & nop & movz & udf & \textit{dst\_x29} & umov \\
		-O3 & Weight & -19.415 & 19.361 & 12.720 & 11.041 & -10.278 \\
		& Feature & \textit{dst\_x29} & dup & sbfiz & movi & dmb \\
		-Os & Weight & 42.171 & -22.245 & -16.490 & -12.939 & -12.103 \\
		& Feature & sdiv & nop & \textit{dst\_x17} & b.cs & \textit{dst\_x25} \\
		\hline
	\end{tabular}
	\vspace{-1em}
\end{table}

\subsection{Evaluation on 32-bit ARM} \label{s:corpus}
We now mirror the experiments in \cref{s:benchmark}, but on a dataset of 32-bit ARM binaries we built and release with this paper.
In addition to having binaries from gcc and Clang and compiled with \{-O0, -O1, -O2, -O3, -Os\}, we compile binaries on versions 6 and 8 for gcc, and 5 and 7 for Clang.
This enables an additional classification task of distinguishing old and new compiler versions.

Moreover, 32-bit ARM is distinctive in its use of conditionally executed instructions, which set the condition flags instead of relying on preceding instructions.
With 64-bit ARM having removed conditionally executed instructions and having nearly double the amount of general-purpose registers at the compiler's discretion,
we aim to determine if DIComP~\cite{dicomp} is also reproducible with high performance on 32-bit ARM.
As before, we also discuss how these ISA differences change the features deemed important by the SVM.

\parhead{Experimental Setup.}
We compile a total of 8,187 binaries on a Raspberry Pi 4 Model B with 2 GB of RAM, varying the compiler family, version, and optimization level.
Our pool of source code includes the GNU coreutils, diffutils, findutils, and inetutils packages, in addition to open-source C and C++ projects that we scraped from GitHub.
Following our case study in \cref{s:comparison}, we remove binaries with duplicate checksums to reduce label noise.

After these filtering steps, the final sample size of our corpus is 6,354 binaries.
In contrast to our 64-bit ARM dataset (\cref{s:benchmark}), we do not apply size filtering because the range of binary sizes is smaller (the largest ARM32 binary is 1.5 MB, compared to 97 MB in our ARM64 dataset).
We perform 10-fold cross validation and report the mean accuracy, runtime for all folds combined, and the five most heavily weighed features for each SVM.

\parhead{Results.}
Our model achieves an accuracy of 0.9920 on the compiler family and 0.7860 on optimization level (with 5 labels), numbers that are consistent with its evaluation on 64-bit ARM binaries.
When the label space for optimization level is reduced, we furthermore observe notable improvements in accuracy.
Firstly, building on the insight that binaries compiled with -O2 and -O3 show high levels of similarity~\cite{Egele14Blanket}, we combine the two flags under one label.
Here, the mean accuracy increases to 0.8560.

Secondly, we further simplify the problem as a binary classification task to detect the presence or absence of optimization, similarly to the first work of Pizzolotto et al.~\cite{Pizzolotto20}.
That is, we treat -O0 as one label, and the other four flags as the second label.
In this setting, we observe an accuracy of 0.9929.
Lastly, our model distinguishes binaries from old and new versions of gcc with 0.9664 accuracy, and of Clang with 0.8522 accuracy.

We also observe runtimes per binary that are consistent with the speed from \cref{s:benchmark}, which we report in \cref{t:arm32}.
Here, the combined training and evaluation time are for all three experiments: compiler family, optimization level, and compiler version.
We note that when the optimization level is treated as a binary classification task, the model speeds up drastically since one vs. rest classifiers are no longer used.

\begin{table}[htb]
	\centering
	\def\arraystretch{1.5}
	\vspace{-0.5em}
	\caption{Accuracies and runtimes for the evaluation of our DIComP-based model~\cite{dicomp} on 32-bit ARM binaries.\label{t:arm32}}
	\footnotesize
	\begin{tabular}{lr|lr}
		\hline
		\textbf{Label} & \textbf{Accuracy} & \textbf{Task per Binary} & \textbf{Runtime} \\
		\hline
		Family & 0.9920 & Extraction & 127.63 ms \\
		gcc Ver. & 0.9664 \\
		Clang Ver. & 0.8522 \\
		Optimization (5) & 0.7860 & Train + Evaluate* & 22.14 ms \\
		Optimization (4) & 0.8560 & Train + Evaluate* & 24.05 ms \\
		Optimization (2) & 0.9929 & Train + Evaluate* & 1.39 ms \\
		\hline
	\end{tabular}
	\vspace{1em}
	
	*The Train + Evaluate tasks correspond to the end-to-end runtime to train all SVMs and predict compiler family, version, and each variation of optimization level classification. (5) indicates all five levels, (4) indicates the case where -O2 and -O3 are grouped, and (2) indicates the case where it is a binary classification task (-O0 against the rest).
\end{table}

\begin{table}[htb]
	\centering
	\def\arraystretch{1.5}
	\caption{Heaviest weights of each SVM and their corresponding features when trained on the 32-bit ARM dataset. Register features are italicized. \label{t:arm32-interp}}
	\footnotesize
	\begin{tabular}{ll|rrrrr}
		\hline
		\textbf{Task} & & \textbf{1} & \textbf{2} & \textbf{3} & \textbf{4} & \textbf{5} \\
		\hline
		Family & Weight & 13.704 & -7.499 & -5.789 & -5.037 & 4.231 \\
		& Feature & bxeq & bxls & \textit{dst\_lr} & \textit{dst\_r3} & blt \\
		\hline
		-O0 & Weight & -4.774 & -3.804 & 3.737 & 3.578 & 3.504 \\
		& Feature & mrc & \textit{src\_r6} & \textit{src\_ip} & \textit{src\_sp} & and \\
		-O1 & Weight & 37.137 & 35.471 & -30.865 & -20.777 & -19.942 \\
		& Feature & andne & blne & ands & nop & clz \\
		-O2 & Weight & 27.711 & 25.517 & -15.429 & 10.016 & -8.143 \\
		& Feature & clz & ands & \textit{dst\_pc} & bcs & bhi \\
		-O3 & Weight & 13.905 & 11.487 & 11.071 & 10.626 & -10.225 \\
		& Feature & \textit{src\_r9} & rsb & bhi & bxeq & push \\
		-Os & Weight & 23.722 & -22.803 & 17.638 & -17.572 & 16.887 \\
		& Feature & push & pop & ldrcc & \textit{dst\_lr} & bgt \\
		\hline
		gcc & Weight & -98.766 & -28.465 & 26.945 & 8.583 & 8.221 \\
		(6, 8) & Feature & bxls & subcs & bxeq & \textit{dst\_fp} & movw \\
		Clang & Weight & 137.185 & -79.410 & -27.634 & -25.791 & -19.940 \\
		(5, 7) & Feature & bmi & ldrbeq & popeq & ldreq & ldrbne \\
		\hline
	\end{tabular}
\end{table}

\parhead{Insights from Model Interpretation.}
Due to space constraints, we show the learned classifier weights for optimization level only for the experiment where we considered all five labels (shown in \cref{t:arm32-interp}).
There, we use the same notation for feature names as in \cref{s:benchmark} and \cref{t:pizzolotto-interp}.

First, we note that some of the most emphasized features for the compiler family are the link register (\texttt{lr}; stores the return address) and \texttt{bx} instructions executed conditionally.
The latter is another hallmark of 32-bit ARM, where \texttt{bx} is the \textit{branch and exchange instruction set} instruction.
Whereas 64-bit ARM has one instruction set, 32-bit ARM supports two instruction sets, ARM and Thumb.
Thumb uses 2- or 4-byte long instructions to shrink code size.
Thus, an explanation could be that gcc and Clang binaries can be distinguished by how willing the compiler backend is to switch instruction sets.

For the one vs. rest classifiers on optimization level, we first note the absence of \texttt{nop} as an important feature in contrast to 64-bit ARM.
Instead, the `hallmark' features seem to vary for each optimization level.
For example, -O0 weighs the intraprocedural call register (\texttt{ip}) and stack pointer heavily from more function calls due to the lack of inlining. -Os has an emphasis on \texttt{push} and \texttt{pop}, which we attribute to the compiler grouping pushes or pops for multiple registers into one instruction, instead of emitting an instruction for one register as early as possible (to reduce code size).

Lastly, for compiler versions, \texttt{bx} instructions are again salient for gcc, hinting at a different preference for emitting ARM vs.
Thumb instructions over its version history.
For Clang, we observe load (\texttt{ldr}) instructions that are conditionally executed being heavily weighed --- possibly an improvement in optimizing smaller if-then-else structures.

\begin{table}[htb]
	\centering
	\def\arraystretch{1.5}
	\caption{Heaviest weights of the SVM classifier trained on the CompCert dataset. Register features are italicized. \label{t:compcert-interp}}
	\footnotesize
	\begin{tabular}{ll|rrrrr}
		\hline
		\textbf{Task} & & \textbf{1} & \textbf{2} & \textbf{3} & \textbf{4} & \textbf{5} \\
		\hline
		CompCert & Weight & -9.852 & 8.319 & -3.474 & 3.049 & -2.976 \\
		& Feature & push & \textit{dst\_ip} & pop & bic & movw \\
		\hline
	\end{tabular}
	\vspace{-1em}
\end{table}

\subsection{Distinguishing Binaries from CompCert} \label{s:compcert}
We turn our focus to formally verified compilers, given the increasing presence of ARM processors in embedded and safety-critical systems.
Hypothetically, an oracle that perfectly and quickly separates binaries originating from a formally-verified compiler would be useful as a scanning tool, such that binaries which pass its test can be guaranteed for correctness (i.e., the machine code exhibits the behavior specified by the original source code exactly).

We expect binaries from a formally verified compiler to manifest major differences in code generation, assuming that the compiler would make very conservative optimizations.
In turn, we expect these differences would make our model a good approximation of such an oracle, given its high performance in \cref{s:benchmark} and \cref{s:corpus}.
We thus refine our model to distinguish binaries from CompCert~\cite{Leroy16CompCert}, a formally verified C compiler, against binaries from gcc or Clang.

\parhead{Experimental Setup.}
We use CompCert to cross-compile binaries for 32-bit ARM.
CompCert does not currently support C++ and is not fully compatible with all C features (e.g., arrays of unknown length).
Thus, for projects that failed to build to completion, we retain individual object files that successfully compiled.
In addition, we add several C-based projects (e.g., the PHP engine).
We compile this updated corpus with -O0 and -O2 on \{CompCert, gcc 6, gcc 8, Clang 7, Clang 9\}.

Next, we filter the corpus as in \cref{s:corpus}, eliminating binaries with duplicate checksums or feature vectors, resulting in a sample size of 1,977.
We run our model on a binary classification task, where we assign one label to binaries from CompCert and one label to all other binaries.

\parhead{Results.}
We observe an accuracy of 0.9934 on this task. Feature extraction took 44.18 ms per binary on our model, and the training and classification step took 7.97 ms per binary.
Akin to our previous experiments, we show the features with heaviest weight in \cref{t:compcert-interp}.
Here, we find a notable correlation between the heaviest weights assigned by the SVM and patterns in assembly generated by CompCert.
The weights of \texttt{push}, \texttt{dst\_ip} and \texttt{pop} seem to result from how CompCert implements function calls, and serve as evidence that the SVM learns features which are contrasting between labels.

Instead of \texttt{push \{fp\}} or the equivalent \texttt{str fp, [sp, \#IMM]} to store the frame pointer on the stack, function prologues in CompCert tend to begin with \texttt{mov ip, sp} followed by a subtraction of \texttt{sp}. In fact, assembly generated by CompCert seems to seldom make use of \texttt{fp}.
CompCert's function epilogues also seem distinct, with an addition to \texttt{sp} followed by a branch to \texttt{lr}, the link register (which holds the return address), instead of \texttt{pop} instructions to \texttt{fp} and \texttt{pc}.

\subsection{Surveying ARM Binaries in the Wild} \label{s:scraping}
We now show the applicability of DIComP's technique~\cite{dicomp} to large corpora of readily-available binaries.
We train our model on the 32-bit ARM dataset from \cref{s:corpus}, but it predicts labels for more than 42,000 binaries of unknown provenance.
Our target platforms are three Linux distributions with support for ARM: CentOS 7 (2019), Raspberry Pi OS released on May 7, 2021, and Ubuntu 20.04.

We take binaries (including shared objects) that were prepackaged in the \texttt{/usr/bin} folder of each distribution, and scrape a large set of binaries from their default package repositories.
We list the sample size for each source in \cref{t:sample-size}.
Our classification task is identical to that in \cref{s:benchmark}, where the label space is \{gcc, Clang\} for the compiler family and \{-O0, -O1, -O2, -O3, -Os\} for the optimization level.

\begin{table}[htb]
	\footnotesize
	\caption{Sample size for each Linux distribution, separated by prepackaged binaries and binaries we scraped from their default repository. \label{t:sample-size}}
	\begin{center}
		\def\arraystretch{1.5}
		\begin{tabular}{c|ccc}
			\hline
			Binaries & CentOS & Raspberry Pi OS & Ubuntu \\
			\hline
			Repositories & 9,887 & 17,678 & 12,372 \\
			Prepackaged & 1,009 & 756 & 565 \\
			\hline
		\end{tabular}
	\end{center}
	\vspace{-1em}
\end{table}

\begin{figure}[htb]
	\includegraphics[trim={0cm 0cm 0cm 0cm}, width=\linewidth]{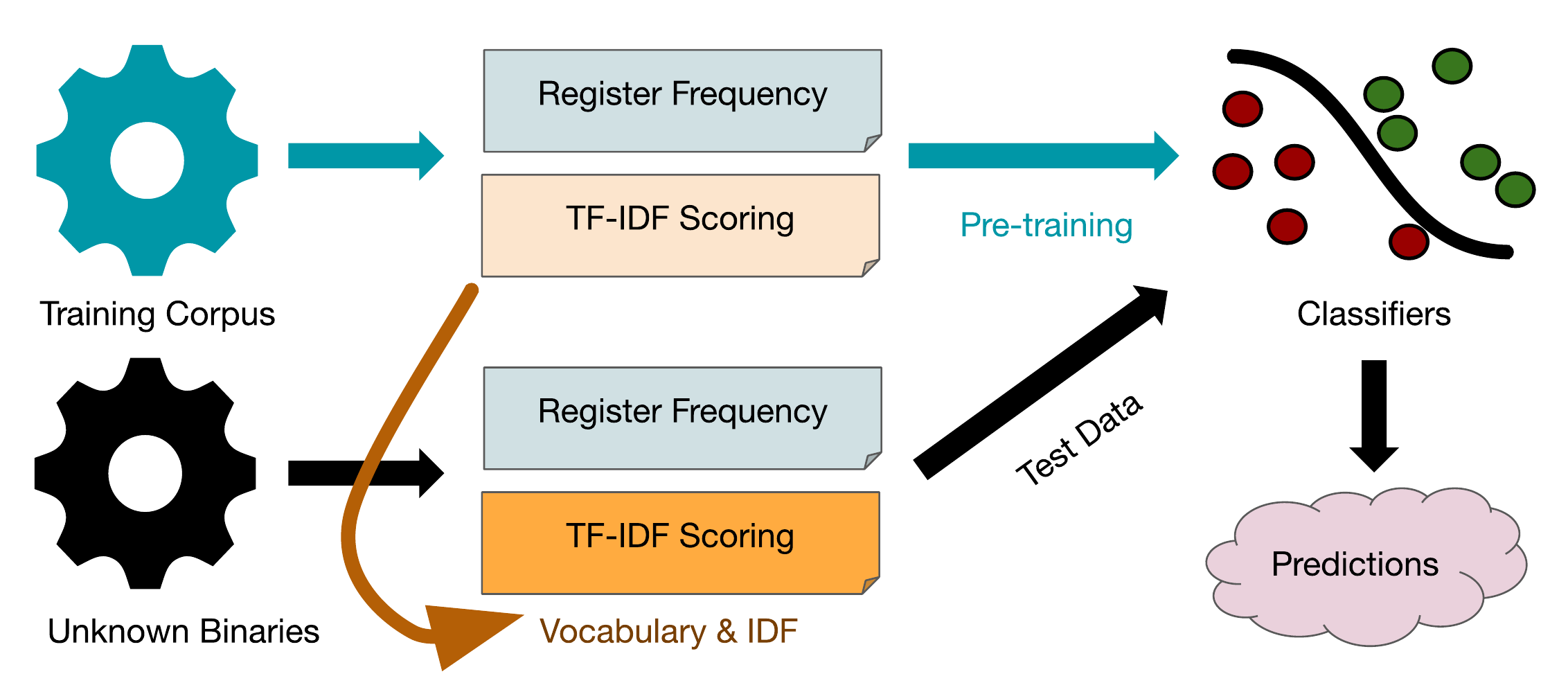}
	\vspace{-1.5em}
	\caption{Overview of our revised model design for generating predictions on binaries without ground-truth labels. We omit the preprocessing step in this figure for brevity.}
	\label{f:variants}
\end{figure}

\parhead{Modifications to TF-IDF Feature Extraction.}
Recall from \cref{s:technical} that we consider TF-IDF scores computed over the vocabulary of distinct opcodes in binaries.
As a result, it is possible for different sets of binaries to have different feature vector lengths and orderings (i.e., different vocabularies).
Furthermore, using the IDF from the unknown binaries would result in a different pattern of emphasis on certain opcodes from that which the SVMs were trained to distinguish.
To address these issues, we save the vocabulary and IDF vectors for all SVMs in the training step.

When extracting features for the unknown binaries, we leave the frequency distributions of register usage unchanged, since it is a statistic profiled per binary (and not the entire corpus being evaluated).
However, for TF-IDF, we calculate the term frequency (TF) of only the opcodes that are present in each SVM's dictionary.
Next, we take the product of each TF vector and the IDF vector paired to the vocabulary that generated the TF vector, and L2-normalize the result.
See \cref{f:variants} for a model overview which reflects the changes for this experiment.

\begin{figure}[htb]
	\includegraphics[trim={0cm 0cm 0cm 0cm}, width=\linewidth]{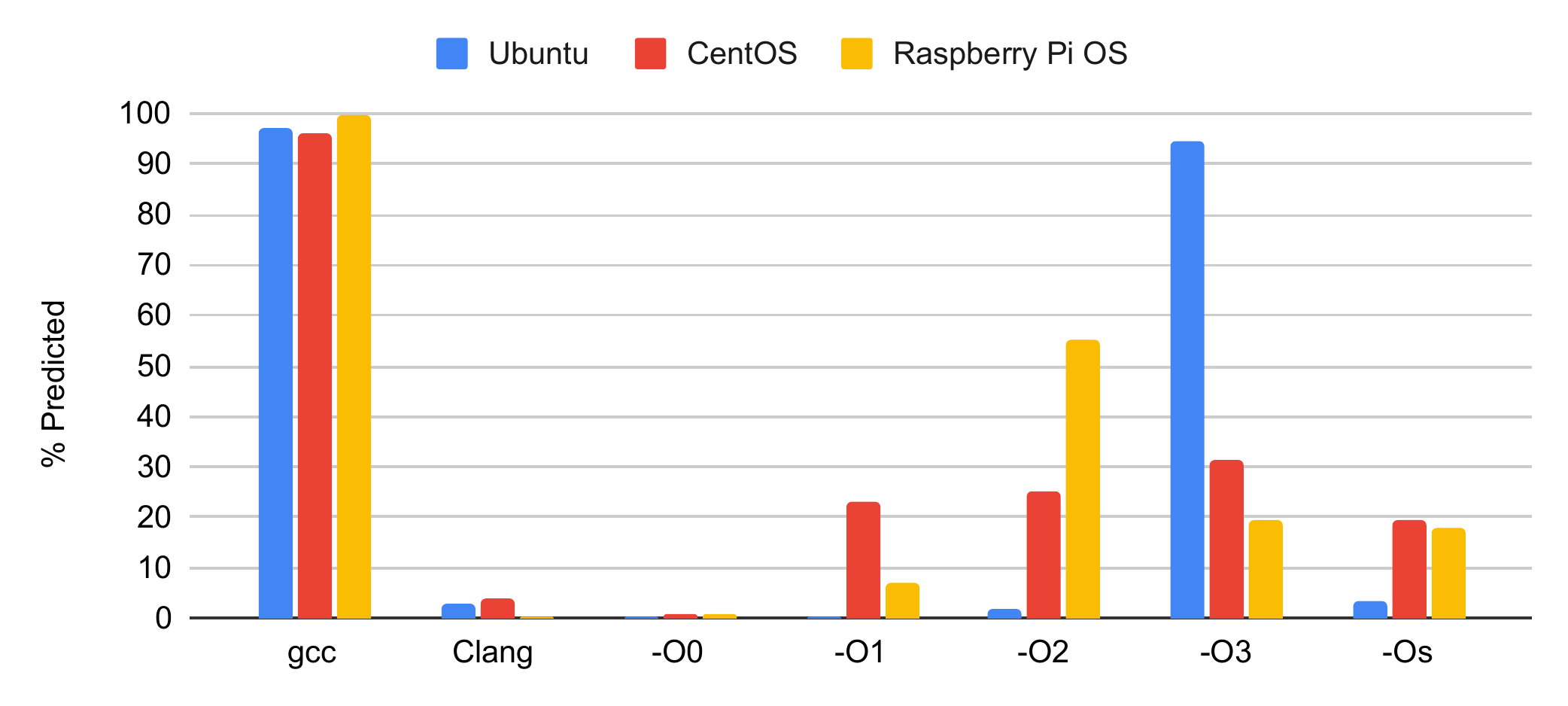}
	\includegraphics[trim={0cm 0cm 0cm 0cm}, width=\linewidth]{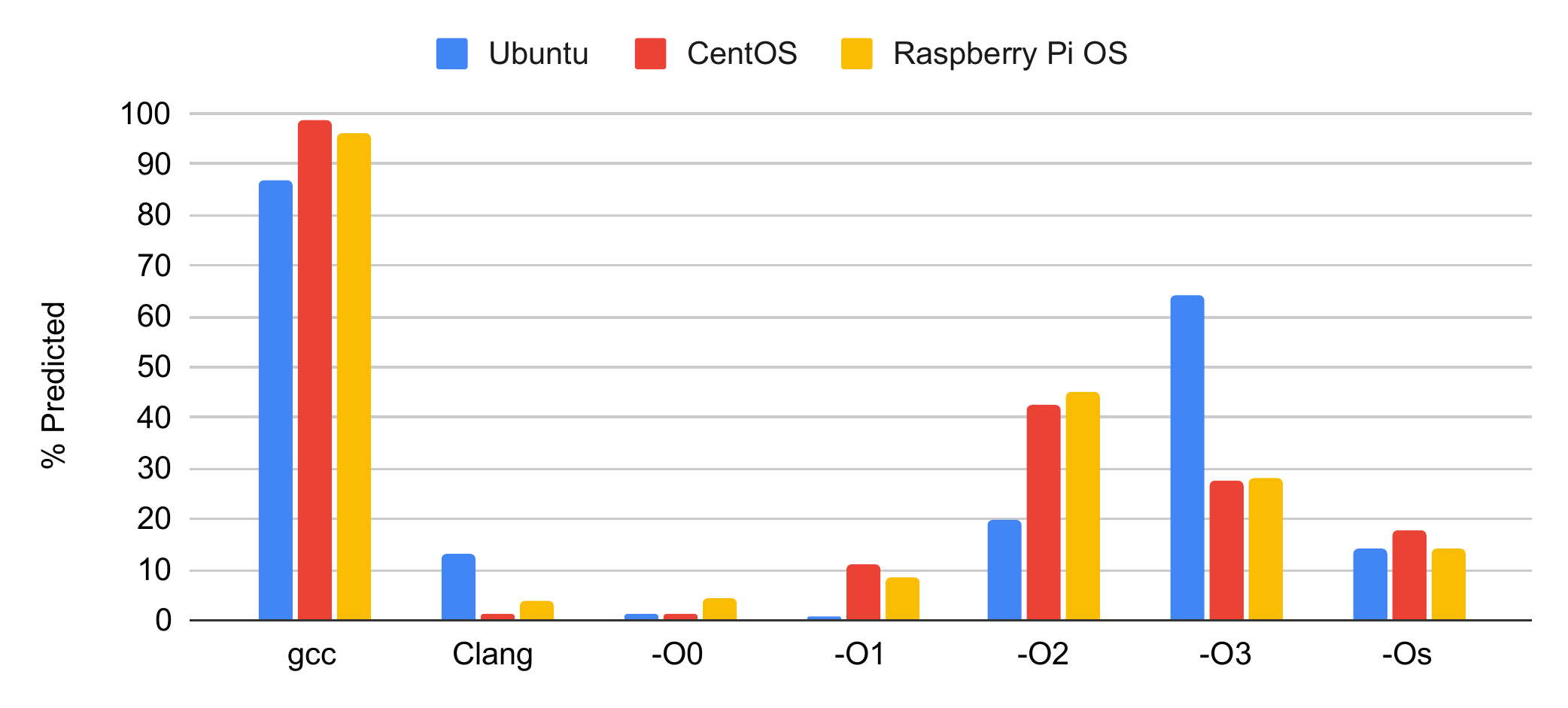}
	\caption{(Top) predictions for binaries prepackaged with each Linux distribution. (Bottom) predictions for binaries scraped from the default package repository of each Linux distribution.}
	\label{f:repos}
\end{figure}

\parhead{Results.}
We show our model's predictions for the compiler family and optimization levels in \cref{f:repos} (Top) for the prepackaged binaries, and \cref{f:repos} (Bottom) for the binaries we scraped from each Linux distribution's package repository.

For both prepackaged and scraped binaries, our model predicts the vast majority of them to have originated from gcc.
In general, our model predicts -O2 or -O3 for most binaries, more specifically an affinity towards -O3 for binaries from Ubuntu, but in contrast -O2 for binaries from Raspberry Pi OS. While the work of Pizzolotto et al. predicts -O2 for most Ubuntu Server 20.04 binaries on x86-64~\cite[Section V-E]{Pizzolotto21}, our dataset is based on Ubuntu for Raspberry Pi (32-bit ARM).

After observing a majority of -Os predictions on macOS Catalina (x86-64), Pizzolotto et al. suggest it may be wrong to assume that unknown binaries were compiled with -O2 when reverse engineering them, despite its popularity in build systems~\cite{Pewny15Xarch}. We note the considerable proportion of -O1, -O3 and -Os in our predictions agrees with their statement.

\section{Threats to Validity}
In this section, we state the assumptions we rely on to build our model and the limitations it can face should these assumptions not hold true. Firstly, we assume the features which were salient for distinguishing different compiler provenances in our work are generalizable to the corpora that the model evaluates. For instance, if our model is deployed as a scanning tool for specialized software built with an in-house toolchain, the assembly code of that software may not exhibit the same idiosyncrasies as the binaries in our datasets.

Secondly, we assume our pool of source code will cause compiler backends to generate substantially different assembly code from each other. However, we acknowledge that this may be less likely for source code that makes extensive use of inline assembly, such as device drivers.

Thirdly, we assume that the binaries our model encounters have one compiler provenance. We recognize this may not be true in some cases, such as a binary which is statically linked with a library produced by a different compiler.

\section{Conclusion} \label{s:conclusion}
We are motivated to design an accurate, fast, and transparent model to recover the compiler provenance of ARM binaries, an area whose research has been limited despite the popularity of ARM devices.
In this paper, we replicate the lightweight approach to feature engineering and model design of DIComP~\cite{dicomp}, adapting it to classify 32- and 64-bit ARM binaries.
We compare our model against a state-of-the-art deep neural network for compiler provenance that is architecture-agnostic~\cite{Pizzolotto21}, first addressing a previously undetected overfitting issue on the work's dataset, and then showing we achieve accuracy on par.
However, we show the lightweight approach trains and evaluates faster by orders of magnitude, and provides insights on features that shape the decision boundary via a direct inspection of model weights.
Finally, we demonstrate a number of case studies using our model, recognizing binaries from a formally verified compiler with near-perfect accuracy, and surveying the provenance of ARM binaries from package repositories.

\bibliographystyle{plainnat}
{\footnotesize
	\setlength{\bibsep}{3pt plus 2pt minus 2pt}
	\bibliography{revisiting-provenance}
}

\end{document}